\documentclass[letterpaper, 10 pt, conference]{IEEEtran}

\IEEEoverridecommandlockouts

\usepackage[T1]{fontenc}
\usepackage{color}
\usepackage{amsmath}
\usepackage{amssymb}
\usepackage{amsfonts}
\usepackage{graphicx}
\usepackage{capt-of}
\usepackage{booktabs}
\usepackage{cite}
\usepackage{url}
\usepackage[hidelinks]{hyperref}
\hypersetup{pdfauthor={Anonymous},pdftitle={Move First, Commit Later},hypertexnames=false}
\usepackage{algorithm}
\usepackage{algorithmic}
\usepackage{enumitem}
\setlist[itemize]{leftmargin=1.2em, itemsep=1pt, topsep=2pt, parsep=0pt}
\let\ralthebibliography\thebibliography
\renewcommand{\thebibliography}[1]{\ralthebibliography{#1}\small}

\title{\LARGE \bf
Move First, Commit Later: Selective LiDAR-to-BIM Global Initialization by Verifying Proposals Against Competing Explanations
}

\author{Yujie Zhang, Yuxuan Guo, Zeheng Fan, Jinyue Liu, Xiaohui Jia, Guang Yang

    \thanks{This work has been submitted to the IEEE for possible publication. Copyright may be transferred without notice, after which this version may no longer be accessible.}

	\thanks{
		Y. Zhang, Y. Guo, Z. Fan, X. Jia, and J. Liu are with the School of Mechanical Engineering, Hebei University of Technology, Tianjin 300401, China. (202421202009@stu.hebut.edu.cn; 202421202013@stu.hebut.edu.cn; 
        202621202043@stu.hebut.edu.cn;
        2010081@hebut.edu.cn; ljy@hebut.edu.cn)

        G.Yang is with the School of Mechanical Engineering, Hebei University of Science and Technology, Shijiazhuang, Hebei, 050018, China. (y\_guang@126.com)
        
        Corresponding authors: X. Jia and G.Yang.
	}

}

\makeatletter
\let\@oldmaketitle\@maketitle
\renewcommand{\@maketitle}{\@oldmaketitle
  \begin{center}
    \includegraphics[width=\textwidth]{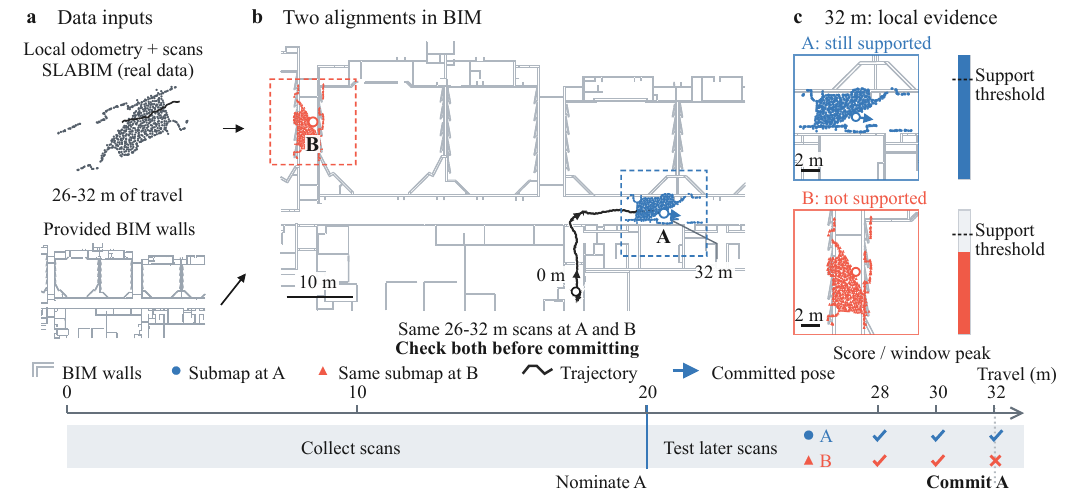}
    \setcounter{figure}{0}
    \captionof{figure}{\textbf{Move first, commit later: deciding when to commit a global alignment.} Real data from the public SLABIM dataset (LiDAR+IMU recordings and the provided BIM). (a) Inputs: local odometry with scans and the BIM walls. (b) The same 26--32 m submap is shown under two global alignments, A and B. (c) These scans support A, whereas B falls below the support threshold. Bottom: A is nominated at 20 m. Both alignments remain supported at 28 and 30 m, so the output is withheld. A is committed at 32 m, when no retained competing alignment remains supported.}\label{fig:teaser}
  \end{center}\vspace{-1.2em}}
\makeatother

\begin{document}

\maketitle
\raggedbottom
\thispagestyle{empty}
\pagestyle{empty}

%%%%%%%%%%%%%%%%%%%%%%%%%%%%%%%%%%%%%%%%%%%%%%%%%%%%%%%%%%%%%%%%%%%%%%%%%%%%%%%%
\begin{abstract}
Global LiDAR-to-BIM initialization must determine both a global alignment and when observations justify committing to it. Repetitive interiors can support several high-scoring alignments, leaving this decision unresolved. We present a decision layer over STSL that nominates and freezes an alignment from submap votes, then checks it alongside actual front-end competitors. Geometric scores assess support on subsequent observations, while cumulative route checks refute explanations that conflict with the building model. The system commits when an eligible submap supports the nominee without supporting a nonequivalent competitor, and otherwise reports the unmet conditions. Across 15 simulated trajectories, it commits correctly in 10 and withholds 5 rotationally ambiguous cases, whereas voting alone makes 8 wrong commits. On 17 real sequences, it commits correctly in 15 and withholds 2, with no wrong commits. These results show that deciding when to commit avoids the wrong initializations observed with fixed-length registration and particle filtering, at the cost of additional travel.

\end{abstract}


%%%%%%%%%%%%%%%%%%%%%%%%%%%%%%%%%%%%%%%%%%%%%%%%%%%%%%%%%%%%%%%%%%%%%%%%%%%%%%%%
\section{Introduction}
\label{sec:introduction}

Building information model (BIM) provides a spatial reference for indoor robot operations. Without a pose prior, global initialization must place the local LiDAR--inertial odometry (LIO) trajectory in this reference. An incorrect placement can associate poses or navigation goals with the wrong location despite consistent local motion. We consider inspection runs that can continue locally while the BIM alignment remains unresolved: the question is not only which alignment is best, but when the evidence is enough to commit it.

Repeated rooms, corridors, and rotationally similar layouts can support several alignments. A registration lead alone may leave these alternatives unresolved, and the observation length needed to distinguish them depends on the route. Figure~\ref{fig:teaser} shows a SLABIM sequence~\cite{huang2025slabim}: the system withholds its output while two alignments remain supported, then commits to alignment~A at 32\,m when later scans distinguish them.

We study commitment as a sequential decision alongside alignment estimation. Nomination fixes an explanation to test; commitment requires sufficient support relative to retained alternatives. This connects to selective prediction, which evaluates accepted outputs together with abstention~\cite{el2010foundations,franc2023optimal}. For localization, withholding can indicate either insufficient nominee support or continued support for another placement; reporting the reason preserves useful evidence for downstream tasks.

Our procedure accumulates submap votes from STSL~\cite{qiao2025speak}. It then fixes the selected alignment and retains competing proposals. Subsequent complete submaps assess these explanations with the same front-end scorer, while cumulative route checks test their physical feasibility. Commitment requires support for the nominee without support for a retained nonequivalent competitor. Otherwise, the report identifies the unmet conditions.

Across 15 simulated trajectories in two layouts, the full procedure commits 10 correct alignments and withholds 5 rotationally ambiguous cases with the room determined and heading unresolved. Pure majority voting and removing the direct competitor veto each produce 7 correct and 8 wrong commits. On 17 real sequences, the full method commits correctly in 15 and withholds 2. On the same sequences, fixed-length registration, particle filtering, and a score-ratio rule over the same candidates make wrong commitments; our method instead waits for distinguishing evidence.

The contributions are:
\begin{itemize}
    \item A formulation separating alignment nomination from first commitment and reporting insufficient evidence or supported competitors.
    \item A comparative verification procedure that tests a fixed STSL nominee alongside front-end alternatives using subsequent geometry and cumulative route feasibility.
    \item Paired simulation ablations and a protocol measuring correct commitment, wrong commitment, withholding, distance, and time, with 17 real sequences, registration and particle-filter comparisons, and threshold sensitivity.
\end{itemize}

\section{Related Work}
\label{sec:related_work}

\subsection{Global LiDAR-to-BIM and scan-to-map registration}

Global LiDAR-to-BIM initialization estimates the rigid transformation between an observed submap and an as-designed model without a pose prior. Repeated structures create perceptual aliasing, also long studied in visual place recognition~\cite{cummins2008fabmap}.

STSL~\cite{qiao2025speak} extracts shared LiDAR--BIM features, generates candidates through Hough voting, and verifies them with an occupancy-aware score on SLABIM~\cite{huang2025slabim}. For measured maps, 3D-BBS~\cite{aoki20243d} uses branch-and-bound point-cloud registration, while learned descriptors retrieve locations~\cite{uy2018pointnetvlad,komorowski2021minkloc3d}. LIO-BIM~\cite{stuehrenberg2025liobim} and BIM-Loc~\cite{zhang2026bimloc} also exploit BIM, and the latter assumes an available initial pose.

\subsection{Sequential and multi-hypothesis localization}

Beyond one-shot candidate generation, another line of work uses sequential observations to maintain or reassess pose hypotheses. Monte Carlo Localization (MCL) and its robust variants rely on particles for global localization and kidnapped-robot recovery~\cite{fox1999monte,thrun2001robust}. Occupancy-grid priors and particle-filter baselines have also been applied to BIM-based localization~\cite{torres2022occupancy}. Reliable-loc~\cite{zou2025reliable} reweights particles with spatially verifiable cues and monitors sequential uncertainty to avoid premature convergence. Active localization deliberately moves the robot to gather observations that reduce ambiguity~\cite{burgard1997active}. Failure detection and integrity monitoring further evaluate whether a pose estimate can be trusted~\cite{akai2022detection,hafez2024safe}.

\subsection{BIM, floorplan, and architectural-graph localization}

Architectural priors offer another route, through semantic and graph-based representations. Specifically, BIM-generated maps support 3D LiDAR localization using persistent walls, corners, and rooms~\cite{yin2023semantic}. iS-Graphs~\cite{shaheer2023graph} aligns an Architectural Graph with an online Situational Graph. When structural symmetry produces multiple optimal matches, the method delays unique association until more observations arrive. A later tightly coupled extension also estimates deviations between as-planned and as-built geometry~\cite{shaheer2025tightly}.

These approaches show that delaying commitment under ambiguity is an established strategy. iS-Graphs delays association when graph matching retains structural symmetry; our method instead checks front-end competitors and route feasibility, without explicit symmetry detection, and reports unmet commitment conditions.

\subsection{Selective decisions and quantified abstention}

Withholding an uncertain estimate is also related to selective prediction, where a model may abstain to reduce errors on the accepted outputs. Selective classification formalizes the trade-off between risk and coverage~\cite{el2010foundations}. Reject-option formulations examine how acceptance decisions affect selective risk~\cite{franc2023optimal}. These works supply decision-theoretic context rather than localization baselines.

\section{Method}
\label{sec:method}

The decision is governed by two questions: whether a new submap supports the nominee without supporting any nonequivalent competitor, and whether the accumulated route remains feasible under each explanation.

\subsection{Problem and Decision Interface}
\label{sec:problem}

We consider global initialization without a pose prior, using LiDAR, inertial measurements, and BIM. The BIM supplies walls, building boundaries, spatial regions, and connections. LiDAR--inertial odometry provides point clouds and a trajectory expressed in the same local frame $L$; $B$ and $R$ denote the BIM and robot frames. At sample $t$, the planar robot pose is ${}^{L}T_{R,t}\in\mathrm{SE}(2)$. A fixed alignment $H={}^BT_L$ places the trajectory in the BIM through
\begin{equation}
{}^{B}T_{R,t}=H\,{}^{L}T_{R,t}.
\label{eq:placement}
\end{equation}
The decision layer determines when to commit this alignment. Ground truth does not enter its online processing.

\begin{figure*}[t]
    \centering
\includegraphics[width=0.9\textwidth]{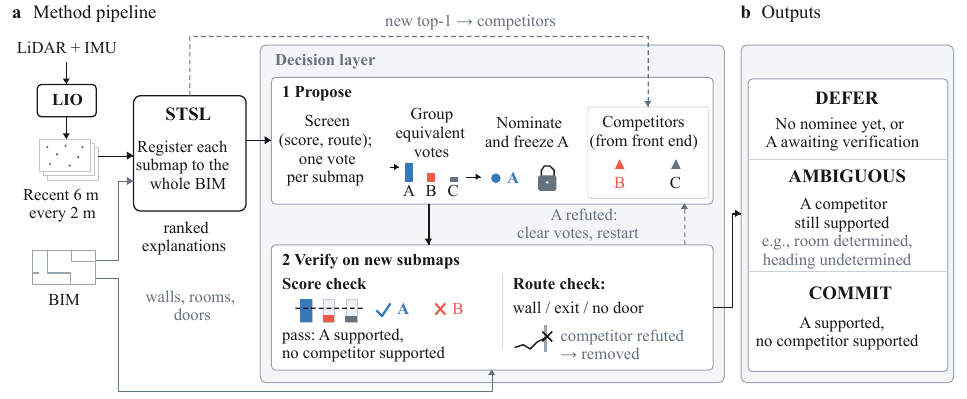}
    \caption{Sequential global initialization. STSL supplies candidate alignments in a common LIO frame. Score-screened, route-feasible candidates contribute at most one vote per window toward nomination. The nominee is fixed and compared with retained front-end competitors on subsequent complete submaps. A new, feasible leading explanation joins the competitor list before eligible-window scoring. A single eligible window supporting the nominee without a supported, nonequivalent competitor commits the fixed alignment; otherwise, the system reports the current evidence and continues observing. Cumulative route checks apply at every window.}
    \label{fig:method-overview}
\end{figure*}

\textsc{Defer} indicates that no alignment has been nominated or that evidence from a new submap is still insufficient to confirm the nominee. \textsc{Ambiguous} takes precedence if the new submap supports a competing explanation that is not equivalent to the nominee; the report lists the supported explanations. \textsc{Commit} indicates that the new submap supports the nominee and no nonequivalent competitor. Figure~\ref{fig:method-overview} outlines the pipeline.

\subsection{Submaps and STSL Candidates}
\label{sec:frontend}

Submap $Q_k$, indexed by $k$, uses target traveled length $\ell>0$ and stride $\Delta d>0$, retaining the latest scans. Voting begins after a complete submap is available. STSL returns $C_k=\{(Y_{k,j},s_{k,j})\}_{j=1}^{K_k}$, where $K_k=|C_k|$ is the output count and $s_{k,j}$ is the native occupancy-aware score. Raw submap points lie in $L$. Let $S_k$ denote the processed planar feature frame. The saved preprocessing transform has planar projection $F_k={}^{S_k}T_L\in\mathrm{SE}(2)$, and $Y_{k,j}={}^BT_{S_k}$. Candidate export and fixed-alignment scoring therefore use
\begin{equation}
H_{k,j}=Y_{k,j}F_k,\qquad Y_k(H)=HF_k^{-1}.
\label{eq:anchor}
\end{equation}
Both operations use the same planar projection of the saved ground preprocessing transform, expressing proposals as alignments from the common LIO frame to the BIM and fixed alignments in each submap's scoring frame.

STSL~\cite{qiao2025speak} forms triangle descriptors from walls and intersecting corners, generates pose hypotheses by hierarchical Pose Hough voting, and evaluates them with an occupancy-aware scorer. We retain its alternatives and reuse this scorer on later submaps.

\subsection{Voting and Nomination}
\label{sec:nomination}

Let $\alpha\in(0,1]$ be the shared relative-score threshold. Within each source window, retain candidates scoring at least $\alpha$ times its positive raw front-end peak. Apply the cumulative route predicate $\Phi_k(H)$ defined in Section~\ref{sec:route}; the highest-scoring feasible candidate contributes a vote. Let $\eta$ mark the earliest source window permitted to vote after a restart, and $\mathcal V$ contain at most one vote per window since $\eta$. Historical candidates must also pass the current route check when votes are assembled. Compare alignments by placing the current robot pose through~\eqref{eq:placement}: equivalence requires absolute $x$ and $y$ differences each strictly below position tolerance $\varepsilon_p$, and wrapped heading difference strictly below angular tolerance $\varepsilon_\theta$. Write $H\sim_k G$ when alignments $H$ and $G$ satisfy these conditions at the current window.

Votes are grouped by complete-link clustering under this equivalence relation, with each group's medoid as its representative. Denote the two largest group counts by $b_1,b_2$, using zero for an absent group, and the required count and lead by $b_{\min},m_{\min}$. If $b_1\geq b_{\min}$ and $b_1-b_2\geq m_{\min}$, nominate the leading medoid $H_\star$; source windows need not be consecutive.

\subsection{Fixed Nominee and Competitor List}
\label{sec:roster}

Freeze $H_\star$ and maintain competitor list $\mathcal R$. Archive $\mathcal A$ retains source-score-screened front-end alignments across restarts. At nomination, admit archived alignments that remain route-feasible and nonequivalent to $H_\star$, retaining exact duplicates only once, regardless of whether they won votes or were rediscovered; only windows since $\eta$ may vote.

At every subsequent window, including ineligible ones, append its raw highest-scoring candidate $h_k$ if route-feasible and nonequivalent to the nominee and every retained competitor. The updated list is scored immediately when eligible, without changing the nominee or votes or imposing another wait.

\subsection{Verification on Subsequent Observations}
\label{sec:verification}

Let $f_N$ be the nomination window's last frame, $a_k,z_k$ the initial and final frames of $Q_k$, and $f_V$ the latest validation endpoint, initialized to $f_N$; eligibility $e_k=(a_k>f_N\ \land\ z_k>f_V)$ ensures that the first complete verification submap lies wholly after nomination and each is evaluated once. Later validation submaps retain length $\ell$ and stride $\Delta d$ and may overlap each other.

After route removals, evaluate $H_\star$ and all remaining competitors on the same eligible submap using~\eqref{eq:anchor}, without re-registration. Let $q_k(H)$ be a fixed-alignment score and $p_k$ the raw current front-end peak, or zero for empty output. Define
\begin{equation}
P_k=\max\!\left(\{0,p_k,q_k(H_\star)\}\cup
\{q_k(H):H\in\mathcal R\}\right).
\label{eq:reference}
\end{equation}
Support requires $P_k>0$ and $q_k(H)\geq\alpha P_k$. The same $\alpha$ governs source screening, nominee support, and competitor support. Every physically retained competitor contributes to the peak, including members currently equivalent to the nominee.

Store supported fixed alignments in $\mathcal S$, the latest eligible window's evidence. Let $\mathcal U_k$ contain the route-feasible competitors in $\mathcal S$ currently nonequivalent to $H_\star$, and let $v_k$ indicate $H_\star\in\mathcal S$ and $\mathcal U_k=\varnothing$. An eligible window satisfying $v_k$ immediately commits $H_\star$. Low-scoring competitors remain for reassessment; unsupported nominees trigger neither timeout nor replacement. Without a new eligible window, retained evidence updates the report using current route feasibility and equivalence, without authorizing a new commitment.

\subsection{Cumulative Route Checks}
\label{sec:route}

The predicate $\Phi_k(H)$ checks the complete accumulated LIO route placed by~\eqref{eq:placement} at every window. FAST-LIO2~\cite{xu2022fast} supplies an append-only trajectory without loop closure. The BIM graph contains spatial regions and located connection channels. Feasibility requires staying inside the building and avoiding walls; a direct transition whose segment is covered by two known regions also requires a registered channel, beyond graph adjacency.

A shared region boundary that the map confirms is wall-free supplies a connection channel. Unannotated interior space leaves region connectivity unresolved while building and wall checks remain active. Doorway centerlines use tolerance $d_{\mathrm{door}}$; square grid cells of width $r$ give a boundary guard $\sqrt{2}\,r$, the cell diagonal. Numerical settings are reported below.

A route contradiction permanently removes a competitor from the attempt. Refuting $H_\star$ clears the nominee and votes and moves $\eta$ to the next window; the archive and accumulated route remain. Later samples preserve contradictions on an unchanged prefix, while different alignments receive their own checks.

\subsection{State Transitions}
\label{sec:states}

Algorithm~\ref{alg:initialization} combines nomination, competitor updates, and commitment. Otherwise, reports identify insufficient votes, unavailable new data, unsupported nominee, or supported competitors. A determined-room, unresolved-heading report requires a supported nominee, all supported explanations placing the current robot in the same known room, and a competitor heading difference of at least $\varepsilon_\theta$.

\begin{algorithm}[tbp]
\footnotesize
\raggedright
\caption{Sequential global initialization}
\label{alg:initialization}
\label{alg:decision} % Compatibility alias for the same Algorithm 1.
\begin{algorithmic}[1]
\REQUIRE LIO trajectory, BIM, submaps $Q_k$, candidates $C_k$, transforms $F_k$
\ENSURE Decision state and, upon \textsc{Commit}, fixed alignment
\STATE Initialize $\mathcal A,\mathcal V,\mathcal R,\mathcal S\gets\varnothing$; $\eta\gets1$; no nominee
\FOR{each complete submap window $k$}
  \STATE Update $\Phi_k$; export candidates by~\eqref{eq:anchor}; identify $h_k$ and archive screened alignments in $\mathcal A$
  \IF{a nominee exists and $\neg\Phi_k(H_\star)$}
    \STATE Discard nominee; clear $\mathcal V,\mathcal R,\mathcal S$; $\eta\gets k+1$; emit \textsc{Defer}; \textbf{continue}
  \ENDIF
  \IF{no nominee exists}
    \STATE Assemble and group $\mathcal V$ from windows $\eta$ through $k$ as in Section~\ref{sec:nomination}
    \IF{$b_1\geq b_{\min}$ and $b_1-b_2\geq m_{\min}$}
      \STATE Freeze winning medoid $H_\star$; $f_N,f_V\gets z_k$; clear $\mathcal S$
      \STATE $\mathcal R\gets\{H\in\mathcal A:\Phi_k(H),\ H\nsim_k H_\star\}$
    \ENDIF
    \STATE Emit \textsc{Defer}; \textbf{continue}
  \ENDIF
  \STATE Remove route-refuted competitors from $\mathcal R,\mathcal S$
  \STATE Append $h_k$ to $\mathcal R$ if it exists, $\Phi_k(h_k)$, and $h_k\nsim_k G$ for every $G\in\{H_\star\}\cup\mathcal R$
  \IF{$a_k>f_N$ and $z_k>f_V$}
    \STATE Score $\{H_\star\}\cup\mathcal R$ at $Y_k(H)=HF_k^{-1}$; compute $P_k$ by~\eqref{eq:reference}
    \STATE $\mathcal S\gets\{H\in\{H_\star\}\cup\mathcal R:P_k>0,\ q_k(H)\geq\alpha P_k\}$; $f_V\gets z_k$
    \STATE Form $\mathcal U_k$ and $v_k$; if $v_k$, emit \textsc{Commit} with $H_\star$ and \textbf{stop}
  \ENDIF
  \STATE Refresh $\mathcal U_k$ from $\mathcal S$; report \textsc{Ambiguous} if $\mathcal U_k\ne\varnothing$, else \textsc{Defer}; attach reasons and any partial report (Section~\ref{sec:states})
\ENDFOR
\end{algorithmic}
\end{algorithm}

\section{Experiments}
\label{sec:experiments}

\subsection{Protocol and Implementation}

\begin{figure}[t]
\centering
\includegraphics[width=\columnwidth]{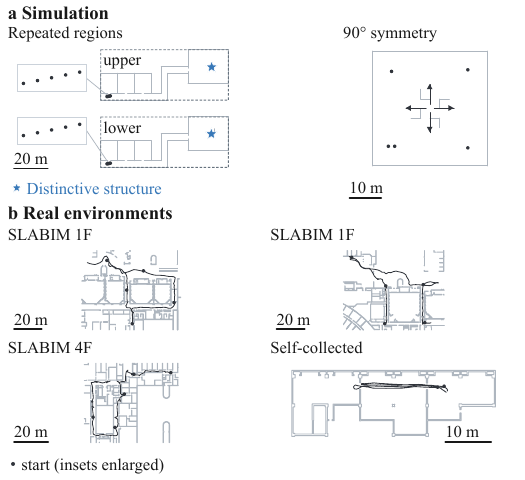}
\caption{Experimental environments. (a) Two repeated regions with five starts each and a strictly $90^\circ$ symmetric room with five starts; stars mark a distinguishing wall and its open-space counterpart, and insets enlarge clustered starts. (b) Three SLABIM regions with five sequences each and the self-collected environment with two sequences; gray lines connect reference-aligned route samples, and dots mark sequence starts.}
\label{fig:environments}
\end{figure}

We evaluate first commitments on 15 simulated trajectories and 17 real sequences. The simulation contains two layouts (Fig.~\ref{fig:environments}a): two repeated regions, traversed from five offset starts in each region, and a strictly $90^\circ$ rotationally symmetric room with five starts. LiDAR and IMU data recorded in NVIDIA Isaac Sim pass through the same LIO, submap, STSL, and decision pipeline as the real sequences. The real data (Fig.~\ref{fig:environments}b) comprise all five supplied sequences in each of the three SLABIM regions~\cite{huang2025slabim}, totaling 15 sequences, and two self-collected sequences. The five sequences within each SLABIM region share the same environment. Each sequence is processed from the beginning of its recording, and travel is measured from the first retained LIO scan.

Map preparation manually unites CAD polygons carrying the same hall label on SLABIM's first floor; this prior serves both first-floor regions. Wall-free shared CAD boundaries are connected automatically. Our method and MCL load identical prepared room, channel, building-boundary, and wall geometries. STSL uses the BIM wall representation, while AMCL uses an occupancy-grid projection.

Unless varied, $\ell=6$\,m, $\Delta d=2$\,m, $\alpha=0.8$, $\varepsilon_p=1.5$\,m per axis, $\varepsilon_\theta=18^\circ$, and $(b_{\min},m_{\min})=(3,2)$. A single passing eligible submap authorizes commitment. The geometric grid has $r=5$\,cm, boundary guard $\sqrt{2}r=5\sqrt{2}$\,cm, and doorway tolerance $d_{\mathrm{door}}=0.20$\,m. Parameters were selected on the simulated trajectories and the two self-collected sequences only. Eleven SLABIM sequences, all of 1F Region3 and the first three in each of the other two regions, were processed during development; the remaining four, the last two in each of 1F Region2 and 4F Region1, were processed once after the configuration was frozen and are part of the 15.

Predictions are sealed before reference evaluation. Correctness requires robot-position error below 1\,m and heading error below $5^\circ$ at the first commitment; withheld runs remain withheld. SLABIM references compose timestamp-matched scan-to-map links with the provided map-to-BIM transform; the self-collected sequences use the existing reference alignments. The planar reference retains the $x,y$ components of the composed three-dimensional body-position reference and computes heading as $\operatorname{atan2}(R_{21}, R_{11})$, where $R$ is the full spatial reference rotation composed with the LIO body rotation before planar projection. We report both commit rate, $(C+W)/n$, and wrong-commit frequency, $W/n$, where $C, W, n$ denote correct commits, wrong commits, and evaluated sequences or seeded filter runs. The latter differs from conditional risk $W/(C+W)$. Travel summaries include committed runs only.

STSL registers all scans accumulated from the start to each fixed length once and returns its first-ranked alignment. We test 6, 10, 15, 20, 30, and 40\,m, the full trajectory offline, and a separate prefix ending at our first commitment. Unreached lengths and unavailable references are excluded. All SLABIM sequences use the same preprocessing. MCL~\cite{fox1999monte,thrun2001robust} uses the same candidates as our method: 353 windows with 353,095 candidates. Its eight configurations combine 2,000/20,000 particles, every first/third input window, and equal/score weighting, each with five seeds. AMCL uses global uniform initialization, 500--20,000 particles, a likelihood-field model, 60 beams, and five seeds per sequence. Its occupancy grid is cut at floor height for SLABIM and at full height for the self-collected maps; on SLABIM, scans are synthesized and leveled using the first complete submap. Its first output is selected by an external largest-cluster weight threshold $w$, initially 0.90.

\subsection{Simulation Ablations}
\label{sec:simulation-results}

Table~\ref{tab:ablation} compares the full method with four variants on unchanged candidate streams. Voting only removes score and route screening and commits at nomination. Nominate and commit retains screening. No competitor veto retains scoring but removes the competitors' ability to withhold output. No route check removes route tests throughout nomination and verification. All other settings, including the voting thresholds, remain fixed.

\begin{table}[t]
\centering
\footnotesize
\caption{Ablation on simulation (15 trajectories) and real data (17 sequences). C/A/W: correct, withheld, wrong. Dist.: mean travel at commit (m); committed sets differ across rows.}
\label{tab:ablation}
\setlength{\tabcolsep}{3.2pt}
\begin{tabular}{lcccc}
\toprule
& \multicolumn{2}{c}{Simulation} & \multicolumn{2}{c}{Real} \\
\cmidrule(lr){2-3}\cmidrule(lr){4-5}
Variant & C/A/W & Dist. & C/A/W & Dist. \\
\midrule
Full method & 10/5/0 & 81.0 & 15/2/0 & 28.0 \\
Voting only & 7/0/8 & 13.9 & 14/3/0 & 18.7 \\
Nominate and commit & 7/0/8 & 13.9 & 16/1/0 & 19.1 \\
No competitor veto & 7/0/8 & 21.9 & 15/2/0 & 26.7 \\
No route check & 6/9/0 & 72.7 & 9/8/0 & 28.2 \\
\bottomrule
\end{tabular}
\end{table}

The full method yields ten correct commits and five withheld runs, all in the rotationally symmetric room, where it reports the room determined and heading unresolved. Voting only, nominate and commit, and no competitor veto each make eight wrong commits. These are the same affected runs: four in the repeated upper region and four in the symmetric room. Competitive verification changes the former to correct commitments and the latter to withholding. Route checks refute four wrong nominees; removing them leaves those four trajectories unresolved. Mean travel over the full method's ten commits is 81.0\,m, with maximum position and heading errors of 0.263\,m and $0.24^\circ$.

\begin{figure}[t]
\centering
\includegraphics[width=\columnwidth]{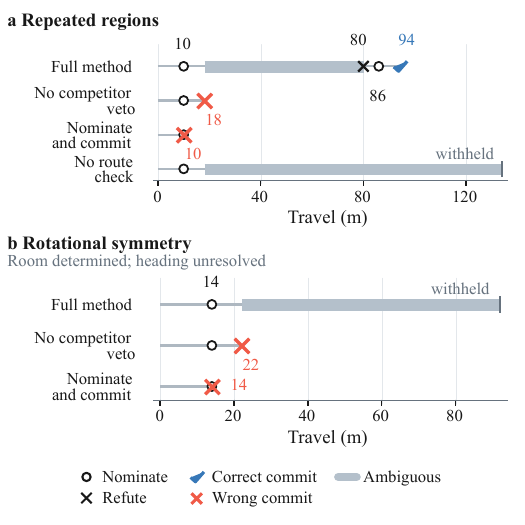}
\caption{Two continuous simulation examples. Open circles mark nomination; black crosses refute a nominee; blue checks and red crosses denote correct and wrong commits. Thick gray spans show recorded \textsc{Ambiguous} outputs. (a) Competition prevents an early wrong commit. A route contradiction at 80\,m enables a new nomination at 86\,m and correct commitment at 94\,m. Without route checks, the run remains unresolved to the last window at 134\,m. (b) Strict rotational ambiguity persists through the last window at 92\,m; the full method reports the room determined and heading unresolved from the first eligible verification at 22\,m. Distances shown are rounded.}
\label{fig:simulation-timeline}
\end{figure}

Figure~\ref{fig:simulation-timeline}(a) traces the mechanism. At 10.03\,m, votes nominate an alignment displaced by 39.94\,m. On the first eligible submap at 18.02\,m, both the nominee and the strongest competitor score 0.926. Supporting the nominee alone would therefore authorize a wrong commitment. The full method waits until its accumulated route contradicts the nominee at 80.03\,m, clears the votes, and later commits correctly. Figure~\ref{fig:simulation-timeline}(b) instead retains competing headings throughout the observed route. Removing score screening, letting verification reuse the nomination data, not adding new top candidates to the competitor list, grouping votes at the start pose instead of the current pose, or removing vote accumulation leaves the simulation outcomes at 10/5/0. In these scenes, the outcome differences come from competitive verification and route checks.

\subsection{Real-Data Comparison and Ablations}
\label{sec:real-results}

Table~\ref{tab:comparison} gives 15 correct commitments and two withheld runs: 13/2/0 on SLABIM and 2/0/0 on the self-collected sequences. All four SLABIM sequences first processed after configuration freezing yield correct commitments. Commit travel has median 24.0\,m, mean 28.0\,m, and range 18.0--46.0\,m. The corresponding elapsed sensor time, measured from the first retained scan, has median 134.2\,s and range 73.5--222.5\,s. Median position and heading errors are 0.225\,m and $0.83^\circ$; their maxima are 0.494\,m and $3.93^\circ$. One withheld sequence nominates at 30.03\,m but receives no supporting eligible submap before its last window at 38.02\,m. The other is only 7.36\,m long and supplies one complete submap and one vote.

\begin{table}[t]
\centering
\footnotesize
\caption{First outputs on real data. Percentages use each row's $n$. Confidence rules use development-selected parameters. Withheld includes abstention or no registration output. STSL excludes unreached lengths and unavailable references. MCL pools eight configurations and five seeds; AMCL uses five seeds per sequence.}
\label{tab:comparison}
\setlength{\tabcolsep}{3pt}
\begin{tabular}{lrrrr}
\toprule
Method & $n$ & Correct & Withheld & Wrong \\
\midrule
Ours & 17 & 88.2\% & 11.8\% & 0.0\% \\
Score-ratio rule & 17 & 47.1\% & 0.0\% & 52.9\% \\
Absolute-score rule & 17 & 0.0\% & 100.0\% & 0.0\% \\
\midrule
STSL, 6\,m & 17 & 35.3\% & 17.6\% & 47.1\% \\
STSL, 10\,m & 16 & 56.3\% & 6.3\% & 37.5\% \\
STSL, 15\,m & 16 & 56.3\% & 0.0\% & 43.8\% \\
STSL, 20\,m & 16 & 75.0\% & 6.3\% & 18.8\% \\
STSL, 30\,m & 16 & 62.5\% & 0.0\% & 37.5\% \\
STSL, 40\,m & 14 & 85.7\% & 0.0\% & 14.3\% \\
STSL, full (offline) & 16 & 87.5\% & 0.0\% & 12.5\% \\
\midrule
MCL, $w=0.90$ & 680 & 24.0\% & 17.6\% & 58.4\% \\
MCL, $w=0.999$ & 680 & 10.7\% & 82.2\% & 7.1\% \\
AMCL, $w=0.90$ & 85 & 0.0\% & 22.4\% & 77.6\% \\
\bottomrule
\end{tabular}
\end{table}

Every tested fixed STSL length contains errors, and full-trajectory registration has two. Relaxing only our planar position threshold to the 3\,m translation tolerance used by STSL~\cite{qiao2025speak}, while retaining the $5^\circ$ heading test, changes two 30\,m outputs from wrong to correct; four remain wrong, and other lengths' counts are unchanged. Both thresholds are evaluated with our planar error metric. With all observations up to our commitment, STSL returns 14 correct and one wrong alignment over the 15 available prefixes. The exception is the sequence in Fig.~\ref{fig:teaser}: its accumulated-scan registration has 44.80\,m position error at 32.03\,m, versus 0.225\,m for our committed alignment. Observation extent alone therefore does not determine success, although one-shot registration can commit earlier in distinguishable scenes.

At $w=0.90$, MCL produces 163 correct, 397 wrong, and 120 withheld outputs; AMCL produces 0, 66, and 19. Increasing MCL's threshold to 0.999 reduces wrong outputs to 48 while leaving only 121 commitments. Conditional on any commitment, median travel is 24.0\,m for MCL and 7.0\,m for AMCL; these include wrong outputs and different committed sets.

Two confidence rules share our candidate stream, native scores, route checks, equivalence tolerances, and development data. Both require equivalent leading explanations over $N$ consecutive windows; the current window must also pass a top-two score-ratio threshold or an absolute-score threshold. The grids use $N\in\{1,2,3,5\}$, ratios $\{1,1.05,1.1,1.2,1.3,1.5,2\}$, and absolute thresholds at the 50th, 75th, 90th, 95th, and 99th percentiles of development-window leading scores. A positive leader with no positive nonequivalent runner-up has infinite ratio; nonpositive leaders fail. Development selection prioritizes zero wrong commits, then the number of correct commits, then median commit travel. It selects ratio 1.05 with $N=1$ and absolute score 0.9785 with $N=5$. Both rules match our 10/5/0 simulation outcome and commit earlier. On SLABIM, the selected ratio rule commits on all 15 sequences, with nine wrong placements having 37.6--161.8\,m position error. The absolute-score rule withholds all 17 real sequences because their leading scores stay below every development threshold. On the four sequences first processed after freezing, our method is correct on all four and the ratio rule is wrong on three.

As a test-set-tuned comparison, we also select parameters directly on SLABIM within the same 28 ratio and 20 absolute-score combinations, recomputing the latter's percentile thresholds from SLABIM scores. Neither rule exceeds nine correct commits without a wrong one, compared with our thirteen. Nine ratio combinations and fifteen absolute-score combinations have zero wrong commits, with five--nine and one--nine correct commits, respectively. The selected ratio setting uses ratio 1 and $N=5$, dropping the score margin and relying on leader stability; the selected absolute-score setting uses the SLABIM median score, 0.5793, with $N=3$. Their correct-commit travel medians are 22.00 and 18.01\,m. These results describe tuning on the test data and are excluded from Table~\ref{tab:comparison} and Figure~\ref{fig:commit-wrong}.

All four real-data ablations in Table~\ref{tab:ablation} also make zero wrong commits. Compared with nominate and commit on their 15 jointly correct sequences, the full method waits a median 8.0\,m and mean 9.6\,m longer, and withholds one additional otherwise correct output. Removing route checks instead reduces correct commitments from 15 to nine because physically inconsistent competitors remain. Route checks remove 1,077 competitor instances over the real sequences and refute no nominee, and per-window top candidates add three competitors.

\subsection{Sensitivity and Runtime}
\label{sec:sensitivity}

\begin{figure}[t]
\centering
\includegraphics[width=\columnwidth]{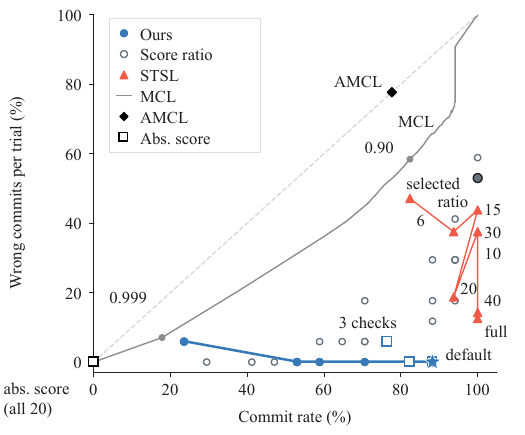}
\caption{Commit rate versus wrong commits per trial; lower right is better. Ours and the confidence rules use 17 sequences per setting; MCL pools 680 runs, AMCL 85, and STSL 14--17 sequences per length (full: offline). The diagonal denotes all commits being wrong. Blue circles sweep $\alpha=0.1$--0.9, squares vary other factors, and the star marks the default. The filled gray circle with a black outline marks development selection among 28 ratio settings. All 20 absolute-score settings coincide at the origin. Test-set-tuned settings are excluded.}
\label{fig:commit-wrong}
\end{figure}

Figure~\ref{fig:commit-wrong} includes all sensitivity settings. On real data, $\alpha=0.1,0.2$ yields 3/13/1 correct/withheld/wrong, 0.3--0.4 yields 9/8/0, 0.5 yields 10/7/0, 0.6 yields 12/5/0, and 0.7--0.9 yields 15/2/0. Mean commit travel at the latter three settings is 30.3, 28.0, and 26.7\,m. Two consecutive passing checks retain 15/2/0 with mean travel 31.5\,m; three yield 12/4/1. Vote thresholds $(2,1),(3,1),(4,2)$ yield 14/3/0; halving and doubling equivalence tolerances yield 15/2/0 and 14/3/0. Default values remain fixed after this sweep. Simulation outcomes stay 10/5/0 for every tested setting.

All sensitivity errors occur on one real sequence, with heading errors of $5.03^\circ$ or $6.66^\circ$ and position errors below 1\,m. Evaluating the same fixed alignment at a later endpoint increases yaw error because planar projection of the tilted spatial reference depends on orientation; these outputs remain counted as wrong.

Timing uses an Intel Core i9-13980HX with 16\,GB of memory; the decision layer is a single-threaded Python implementation. STSL registration takes 0.682\,s per submap on average (0.419--1.221\,s over 316 SLABIM submaps). Over the 17 real sequences, the median per-sequence mean window time of the decision layer is 333.9\,ms (8.7--993.1\,ms). The slowest window, 6.49\,s, occurs at nomination in one 1F Region3 sequence and is spent almost entirely on route checks while the competitor list is built; in that sequence, every other stage stays below 0.7\,s. Consecutive windows arrive 10.75\,s apart in median and at least 5.50\,s apart. LIO, submap assembly, feature extraction, and descriptor preparation are not included.

\section{Limitations}
Commitment is relative to finite STSL proposals and empirical score support, with ambiguity assessed through retained alternatives. Scores express geometric support rather than calibrated correctness probabilities; unproposed alignments remain untested. Parameters were selected on the simulated and self-collected environments, and the four SLABIM sequences processed after freezing form a small held-out check; a larger unseen set would strengthen it. Broader scene diversity remains an extension. References compose existing alignment links rather than independent survey truth. Confirming with a single window can be misled by transient structures absent from the BIM, such as temporary partitions; requiring consecutive confirmations trades travel for robustness to such effects.

Route checks depend on BIM fidelity and LIO coordinates; grid allowances address discretization, while drift and residual spatial tilt remain relevant. The AMCL result reflects initialization and tracking together. End-to-end timing requires synchronized sensing, LIO, preprocessing, and front-end execution; motion selection and post-initialization tracking remain downstream tasks.

\section{Conclusion}
In this paper, we presented a LiDAR-to-BIM commitment layer that fixes a nominee and checks it against front-end competitors on later observations and the accumulated route. In simulation, competitive verification turns wrong commits into correct commits or withholding. On 17 real sequences, it commits correctly in 15 and withholds 2 with no wrong commitment, whereas fixed-length registration, particle filtering, and a score-ratio rule over the same candidates make wrong commitments; this comes at the cost of additional travel.
\bibliographystyle{IEEEtran}
\bibliography{references}

\end{document}